\documentclass[smallextended]{svjour3}       
\smartqed  
\usepackage{graphicx}
\usepackage{hhline}

\usepackage{multicol}        
\usepackage[bottom]{footmisc}

\usepackage[ruled,vlined,linesnumbered]{algorithm2e}
\SetKwRepeat{Do}{do}{while}%

\usepackage{array,multirow,makecell}
\setcellgapes{1pt}

\usepackage{epsfig}
\usepackage{epstopdf}

\usepackage{lipsum,graphicx,multicol}

\usepackage{amssymb}
\usepackage{amsmath}
\makegapedcells
\newcolumntype{R}[1]{>{\raggedleft\arraybackslash }b{#1}}
\newcolumntype{L}[1]{>{\raggedright\arraybackslash }b{#1}}
\newcolumntype{C}[1]{>{\centering\arraybackslash }b{#1}}
\usepackage[justification=centering]{caption}

\usepackage{xcolor}
\begin{document}

%

\title{A General Framework for Complex Network-Based Image Segmentation.}



\author{Youssef Mourchid         \and
        Mohammed El Hassouni \and
        Hocine Cherifi
}


\institute{Youssef Mourchid \at
              LRIT - CNRST URAC 29, Rabat IT Center, Faculty of Sciences, Mohammed V University in Rabat, Morocco\\
              \email{youssefmour@gmail.com}           
           \and
           Mohammed El Hassouni \at
              LRIT - CNRST URAC 29, Rabat IT Center, FLSH, Mohammed V University in Rabat, Morocco. LRIT - CNRST URAC 29, Rabat IT Center, Faculty of Sciences, Mohammed V University in Rabat, Morocco.\\
              \email{mohamed.elhassouni@gmail.com} 
              \and              
              Hocine Cherifi \at 
              LE2I UMR 6306 CNRS, University of Burgundy, Dijon, France.\\
              \email{hocine.cherifi@gmail.com} 
}

\date{Received: date / Accepted: date}

\maketitle

\begin{abstract}
With the recent advances in complex networks theory, graph-based techniques for image segmentation has attracted great attention recently. In order to segment the image into meaningful connected components, this paper proposes an image segmentation general framework using complex networks based community detection algorithms. If we consider regions as communities, using community detection algorithms directly can lead to an over-segmented image. To address this problem, we start by splitting the image into small regions using an initial segmentation. The obtained regions are used for building the complex network. To produce meaningful connected components and detect homogeneous communities, some combinations of color and texture based features are employed in order to quantify the regions similarities. To sum up, the network of regions is constructed adaptively to avoid many small regions in the image, and then, community detection algorithms are applied on the resulting adaptive similarity matrix to obtain the final segmented image. Experiments are conducted on Berkeley Segmentation Dataset and four of the most influential community detection algorithms are tested. Experimental results have shown that the proposed general framework increases the segmentation performances compared to some existing methods.

\end{abstract}

\maketitle

\section{Introduction}

Image segmentation is a fundamental problem in computer vision. It refers to split the image into uniform and homogeneous regions which correspond to meaningful parts of the image. Image segmentation application area varies from the identification of objects from remote sensing data \cite{1,30,31} to the detection of cancerous cells. For example, it can be used to diagnose medical imaging \cite{2}, or to extract interest points for images for identifying the local features of an image \cite{32}. In the literature, a lot of image segmentation algorithms have been proposed. They can be classified into two main categories: Edge detection and Region-based approaches. Edge detection methods are based on the use of discontinuities to detect the edge for segmenting an image. Several methods \cite{43,44} are proposed in this category, which almost are based on the abrupt changes in image intensity or color. In Region-based approaches which is another popular category of segmentation methods, the segmentation is achieved using an iterative manner, until some uniformity criteria are satisfied. The principal methods in this category are based on thresholding \cite{46}, regional growth \cite{48} and graph-based \cite{42}. Graph-based techniques represent the image components into mathematically sound structures which makes the segmentation problem easier and the computation faster and efficient. The problem of the graph based image segmentation methods concerns the partition into several sub-graphs, such that each sub-graph represents a meaningful object of interest in the image. The idea of using graphs as an approach for segmentation was brought by Wu \textit{et al.} \cite{3}. From then on, the study of optimization techniques on the graph attracted much research attention \cite{4,5,6,7,10}. Our work is in this line. In this paper, we propose a framework that uses a weighted region adjacency graph to represent an image as a network, where regions represent nodes in the network. An edge between two nodes is considered if they are adjacent and then the weight is calculated using images features (color and texture). 

Because networks are growing exponentially in size, variety, and complexity, newer and different types of communication networks are emerging, such as social networks, biological network and so on. Most of them are structured into communities, which represent groups of nodes that are heavily connected among themselves but sparsely connected with nodes in other groups \cite{31}. Taking into account the importance of community detection, it is not surprising that many community detection methods have been developed using techniques from different disciplines. So, it is possible to consider those methods for identification of objects in an image. More specifically, an image can be mapped as a graph and community detection approaches can be considered to identify the regions, which correspond to communities in networks. However, using them only has a practical limitation. 

Motivated by graph-based methods and the application of community detection algorithms in graphs, we propose a general framework for image segmentation based on community detection algorithms. In many existing works, complex networks have been formulated as non oriented unweighted networks to simplify the analysis and the computation. Nevertheless, these networks can simplify computation and analysis, but they lost some important information, which affects the network performance. Unlike these existing works, the proposed general framework starts with an initial segmentation of an image in order to build a Region Adjacency Graph (RAG) \cite{33}. The nodes are the initial regions and an edge exists between two nodes if they are adjacent. Then the edges are weighted according to the similarity between meaningful visual features of the regions (texture, color). A community detection algorithm is applied next on the Weighted RAG (WRAG) in order to partition the network into a set of communities. These communities are used to group similar adjacent regions in the image. Indeed, all the nodes that belong to the same community are considered to belong to the same region and merged into a single region in the image. The process is iterated until there is no difference between the uncovered community structures of two successive iterations 


The main contributions of this work are the following:
\begin{itemize}

\item First, we propose to model an image with a Weighted RAG that takes advantages of the topological and visual properties of images (texture, color).

\item Second, we use community detection algorithms developed in the complex networks paradigm in order to solve the segmentation problem.

\item Third, an iterative process is proposed in order to avoid over-segmentation issues.

\end{itemize}


The rest of the paper is organized as follows. Section 2 presents previous investigations related to the proposed general framework. In section 3, we review the necessary background on evaluation metrics used to asses the proposed framework. Section 4 presents the definition of community detection concept. The details of each step of the proposed framework are presented in section 5. Section 6 gives a description of data and methods used for experiments. We reported our experiments results in Section 7. Finally, in section 8, we present our conclusions.

\section{Related work}

Several graph-based techniques are used to segment an image \cite{3,4,5}, a lot of them are based on the optimization of a cost function. In \cite{3}, authors considered a minimum cut criterion, which seeks to partition a graph into a number of subgraphs by minimizing the sum of the edge weights. \\
Felzenszwalb \textit{et al.} \cite{4} proposed an approach derived from a pairwise region comparison to segment the image. It defines a criterion to evaluate if an edge exists between two regions based on an iterative strategy and using a graph representation of the image, to obtain the final segmentation. Their proposed method is almost linear-time in the number of graph edges and can be employed to segment large images. \\
Shi \textit{et al.} \cite{5} used a technique called normalized cut. Authors represent the points to be clustered by an undirected graph, where the nodes represent the points to be segmented and each edge weight represents the similarity between two points. The graph cut is measured by the weights of the total connection from vertices in a set A to all the vertices in the graph, the weight is computed next by measuring a certain image quantity (e.g., color, intensity, etc.) between the two vertices connected by that edge. The Ncut measure tries to minimize the cut and to penalize partitions at the same time in which one set of nodes is only loosely connected to the graph at large. \\
Li \textit{et al.} propose an image segmentation algorithm \cite{6} based on graph modularity optimization. First, they start their algorithm by an initial segmentation using the superpixels method which oversegment the image into a set of small segments. Each segment represents a region in the graph, and then the graph of regions is constructed using two image features to compute the similarity between regions, this similarity is assigned as a weight to the network. Finally, from the regions graph, they apply a community detection algorithm based on modularity optimization.\\
In Abin \textit{et al.} \cite{7}, authors start by an initial segmentation using meanshift and then they construct the network using the similarity between two regions using only color information. This similarity is used as a weight to the edges of regions network. Finally, they apply a community detection algorithm to obtain the segmented image. Nevertheless, the authors didn't use an iterative process to avoid the over-segmentation problem, instead, they use a post-processing algorithm to merge regions with areas smaller than a predefined threshold with other regions. If a region area is smaller than the threshold t, it's merged to the most similar adjacent region in the network.

Linares \textit{et al.} \cite{10} proposed also an algorithm based on community detection algorithms. They start first by an initial segmentation using superpixels method, and then, they consider pixels on the image as nodes on the graph. To construct the network, they use the CIELAB feature to compute the similarity between superpixels in the image. A connection between two nodes is considered only if the weight is smaller than a threshold t. Finally, when the graph of superpixels is constructed, they apply the fast greedy algorithm which is a hierarchical agglomerative algorithm for community detection. However, the proposed approach in \cite{10} uses an unweighted network instead of a weighted network which means that some important informations of the image are lost, which affects the network performance.

The cited graph-based segmentation approach are generally sensitive to noise, and use either the color or the texture as a measure to compute the similarity between image regions, which leads to over-segmentation by neglecting the regularities inside the image. Moreover, most of these methods are based on community detection algorithms that have a high computational cost. To overcome these limitations, the proposed framework starts with an initial segmentation to split the image into regions which should be coherent and preserves most of the information necessary for segmentation. Then, a RAG is used to represent the image where each region represents a node in the graph. An edge between two regions is considered if they are adjacent. In order to weight the RAG, a combination of texture and color features is employed to measure the similarity between nodes. Finally, based on efficient community detection algorithms which strike the best balance between the computational cost and segmentation performance, we extract communities that represent regions in the image. The process is repeated iteratively until the optimal segmentation is achieved.

\section{Background}

As we mentioned previously, the proposed framework starts by an initial segmentation to split the image into small regions, then a RAG is constructed using the adjacency relationship. This RAG is weighted using the similarity between regions based on image features (color, texture). Finally, community detection algorithms are employed for partitioning the weighted RAG into communities to obtain the optimal segmentation in the image. In this section, we describe methods and measures used in the proposed framework.

\subsection{Initial Segmentation algorithms}

The goal of the initial segmentation is to split the image into homogeneous, possibly small regions. Several low-level segmentation methods can be used in this step, such as super-pixel, Meanshift, levelset, and watershed. 

\subsubsection{Super-pixels Algorithm}

The \textbf{super-pixels} algorithm splits the same perceptual region in a multitude of smaller regions. It's usually used as an initial segmentation process for reducing the pixels number and the computational complexity of subsequent tasks. Several studies exist for the extraction of super-pixels. In \cite{8}, authors propose an efficient technique that yields quasi-uniform super-pixels with low computational cost. The results obtained by the method show its efficiency in terms of computational costs and compactness of segments and over-segmentation errors. Authors in \cite{49} use a connected K-means algorithm with convexity constraint for extracting super-pixels. According to a number of regions desired by the user, the image is splited into rectangular regions (segments) using a regular grid. Then, by minimizing a cost function:

\begin{equation*}
       C_{x,y}(i)= \lambda_1.|I(x,y)-I_i|+\lambda_2.|(x-C_{x}^i)^2+(y-C_{y}^i)^2)|
\end{equation*}

Where x and y are the positions of the pixel tested among different segments; $\lambda_1$ and $\lambda_2$ correspond respectively, to the weighting of intensity similarity and convexity constraints; $I_i$ denotes the mean intensity of the $i^{th}$ segment and $C_{x}^i$ and $C_{y}^i$ are the center positions of the $i^{th}$ segment. The super-pixels algorithm tests pixels at the over segment boundaries and assigned them to the new segments. In the proposed framework, a publicly available code \cite{47} is used to get the superpixel initialization.

\subsubsection{Meanshift Algorithm}
The \textbf{meanshift} is a non-parametric iterative algorithm \cite{9} that can be used for a lot of purposes like finding modes, clustering etc. One advantage of Meanshift over other pre-segmentation techniques is that we don't have to specify the number of segments (clusters) because the algorithm itself finds the best number of clusters for the image. To start the MeanShift algorithm on a set of data points X (pixels in the image), we need:

\begin{itemize}
    \item A function N(x) which determines what are the neighbors of a point $x \in X$. The neighboring points are the points within a certain distance. The distance metric is usually Euclidean Distance.
   \item A kernel function K(d) to use in Meanshift. K is usually a Gaussian Kernel, and d is the distance between two data points.
\end{itemize}

Now with the above functions, the process of Meanshift for a set of data points X follows these steps:

\begin{enumerate}
    \item For each data point $x \in X$, find the neighboring points N(x) of x.
    \item For each data point $x \in X$, calculate the mean shift m(x) from this equation:

    \begin{equation*}
        m(x)=\frac{\Sigma_{x_i\in N(x)}K(x_i-x)x_i}{\Sigma_{x_i\in N(x)}K(x_i-x)}
    \end{equation*}
    
    \item For each data point $x \in X$, update $x\gets m(x)$.
    
    \item Repeat 1. for n iterations or until the points are almost not moving or not moving.
    
\end{enumerate}

\subsection{Similarity measures}

Several measures can be employed to compute the similarity for the proposed features of the framework. 

\subsubsection{Color Similarity}

Color in segmentation is an important and straightforward feature. Each pixel in a color image is represented by a three-dimensional vector. We assume that the pixel intensity value of a given region follows a Gaussian distribution. Therefore, the distribution of a region $R_i$ is given by: $R_i \sim N(\mu_i;var_i)$ , where $\mu_i$ is the mean vector of the the pixel intensity computed in the three-dimensional color space in regions $R_i$, $var_i$ denotes the variance of $R_i$.

To measure the similarity between two distributions, various distance measures are proposed in the literature such as the Earth Mover's Distance (EMD) \cite{14}, Kullback-Leibler (KL) Divergence \cite{13}, Mean Distance (MD), etc. We choose to use the Mean Distance (MD) because it is generally a good approximation of the Earth Mover's Distance with a lower complexity. MD can be defined by the formula below:

\begin{equation}
{\rm D_{MD}(R_i;R_j)= (\mu_i-\mu_j)^T (\mu_i-\mu_j)}  
\end{equation}

To transform the color feature distribution distance to a similarity measure, we use a radial basis function kernel:

\begin{equation}
{\rm c_{ij}= exp \left( \frac{- D_{MD}(R_i;R_j)}{2\sigma^2} \right)} 
\end{equation}

where $\sigma$ is a parameter defined by the user.




 

Various color space can be used such as RGB,
YUV, L*a*b, HSV, etc. Choosing an appropriate color space to segment a color image is a crucial step in order to achieve a better segmentation performance. Thanks to its accordance with the human visual system \cite{30}, we choose the LAB color space. It's a 3-axis color-opponent space with dimension L for lightness and A and B for the color opponent dimensions.



\subsubsection{Texture Similarity}

Using only the color feature in the image cannot achieve a good segmentation result, because the color feature in some homogeneous object will decompose image regularities into different segments. Therefore, we propose a texture feature as a solution to remedy this problem. Many recent approaches use wavelet as features \cite{11}, other methods, such as \cite{12}, learn dictionaries of local structures from training images. In this work, we use a feature called Histogram of Oriented Gradients (HOG) which is well known in image processing and computer vision for detecting objects in the image. It computes the number of gradient orientation occurrences in localized parts in the image.
To construct the Histogram of Oriented Gradients we proceed with the following steps: 

\begin{itemize}

\item We need first to calculate the horizontal and vertical gradients; after all, we want to calculate the histogram of gradients for a region image. 

\item In the second step, the image region is 
splitted into small cells of size $C\times C$ 
pixels (C = 8). For each cell, the histogram of gradient directions is computed. The histogram is essentially a vector of 9 bins.

\item In the third step, we use a method called Block normalization to group individual cells into blocks and normalize them to ensure invariance to illumination changes. A Block is represented by a $2\times2$ cells so that each block has a size $2C\times2C$ of pixels (4 histograms).

\item Finally, the final feature vector is calculated for the entire region $R_i$, where the histogram of gradient vectors of blocks $h_c$ are grouped into a single HOG feature vector $H_i$: 
\end{itemize}
\begin{equation}
    H_i=[h_1,...,h_c]
\end{equation}
Where $h_c$ denotes the histogram  of gradient vectors of a block, c is the number of blocks inside a region $R_i$. To compute the similarity between two regions $R_i$ and $R_j$, we use the cosine similarity measure as defined by the formula below:

\begin{equation}
{\rm t_{ij}= cos(H_i,H_j)=   \frac{H_{i}^{T}H_j}{\|H_{i}\|.\|H_{j}\|} }
\end{equation}

where $||.||$ denotes the $L_2$ norm, $H_i$, $H_j$ are respectively the HOG vectors of the regions $R_i$ and $R_j$.

\subsection{Community detection}

Community structure is a property of complex networks, which can be described as the gathering of nodes into communities such that there is a higher density of edges within communities than between communities themselves \cite{34}. The identification of communities is quite useful because nodes belonging to the same community are more likely to share properties. Many algorithms have been proposed for extracting the community structure in networks. Newman \cite{35} has defined a measure called the modularity, which is widely used for evaluating a partitioning of a network into communities:

\begin{equation}
 Q= \Sigma(e_{ii}-a_{i}^2)
\end{equation}

Where $e_{ii}$ denotes the fraction of network edges which are inserted into a community $i$, and $a_{i}^2$ denotes the fraction considering that edges are inserted randomly. The modularity value Q is between 0 and 1. A high value of the modularity means a strong community structure of the network. Another quality measure called stability Qs was introduced in \cite{36} based on the clustered auto-covariance of a dynamic Markov process, which also measures the quality of a partition as a community structure. In order to choose the appropriate community detection algorithms for the built weighted region adjacency graph to extract communities, synthesis papers are used to find and then to assess community detection algorithms. From \cite{17}, \cite{18}, \cite{19}, the algorithms proposed by  Ronhovde and Nussinov \cite{20}, Infomap \cite{21}, Fast greedy modularity optimization algorithm \cite{22} and Louvain \cite{23} are judged to be able for delivering a reasonable estimator of the number of communities for different size of networks and then, outperforms all the state of the art algorithms for detecting communities. Here, we present a brief description of each algorithm.

\subsubsection{Fast multi-scale community detection algorithm using the criterion from Ronhovde and Nussinov (FMCDRN)} 


The algorithm is an improvement of the algorithm in \cite{20} which is based on the minimization of the Hamiltonian of a Potts-like spin model, where the spin state denotes the belonging of the node in the community. To cover multiple community scales, from very small to very large, a resolution parameter is used. To identify relevant scales, the algorithm checks for each given value of the resolution parameter, the stability of the obtained partitions. This is done when we compute for the same resolution parameter, the similarity of partitions obtained, but by starting from different initial conditions. Peaks in the similarity spectrum represent relevant partitions. The algorithm is rather fast and its computation complexity is slightly superlinear in the number of edges of the graph. We refer to the method as RN in the next sections. The aim of the proposed framework is speed efficiency. To deal with that a greedy approach is used which exploits all the available information (i.e. input data and information computed as the algorithm runs).

\subsubsection{Infomap}

The algorithm \cite{21} is based on a compression technique to define the information flow on networks. The algorithm uses random walks of a given length with a given probability of jumping for performing. Walks are considered as sequences of steps inside the community which are followed by a jump through a two-levels nomenclature based on Huffman coding. The two-levels are used to identify nodes in the community and communities from the network. The algorithm uses a coding strategy, where each node codeword is derived from the visit node frequency of an infinitely long random walk. This strategy leads to a compact representation of the walks. Authors showed that the optimal partitioning problem treated as finding the minimum description length for all the walks.
 

\subsubsection{Fast greedy modularity optimization algorithm (FGMDO)}

The algorithm \cite{22} represents the fast version of a previous method proposed by Newman. It starts from a set of nodes that are initially isolated and added edges between them to construct the original graph iteratively. It produces at each step the greatest possible increase of the modularity value of Newman and Girvan. First, the algorithm starts with a number of communities N, each community contains a single node, the communities are repeatedly grouped together iteratively at each step, by choosing the set that results in the largest increase (or smallest decrease) in modularity value. The algorithm runs far more quickly. For networks that have a hierarchical structure with communities at many scales and sparse networks, the algorithm has essentially linear running time. This is not only an advanced technique but it's a technique that has substantial practical implications, as it allows to study networks with a large number of nodes.

\subsubsection{Louvain}

The algorithm finds partitions of large graphs with high modularity value in short time and unfolds a complete hierarchical community structure for the graph \cite{23}. It's divided into two steps which are iteratively repeated. First, the algorithm starts with a weighted graph that contains N nodes and we assign to each community of the network one node. For each node i, the algorithm considers the neighbors j of i, and evaluates the gain of modularity when i and j are grouped into the same community. The node i is then moved to the community for which this gain is maximum, but only if this gain is positive. If no positive gain is possible, the node i stays in its current community. The process of grouping is applied iteratively for all nodes until no further maximization of the gain can be achieved. In the second step of the algorithm a new graph is constructed, whose nodes are now the communities found during the first step. Louvain offers a fair balance between the accuracy of the estimate of the modularity maximum and computational complexity. The output of the algorithm, therefore, provides several partitions. The partition found after the first step contains many communities that contain a small number of nodes. At subsequent steps, larger communities are found due to the iterative grouping mechanism. This process naturally leads to hierarchical decomposition of the graph.

\section{The Proposed Framework}

Due to the inherent properties of the image, segmentation and community detection problems are different. Using community detection algorithms only to segment an image by considering pixels as nodes on the graph, can leads to low performances. The failure of such method can be explained by several reasons. First, when we segment an image, pixels can have different properties, for example, different colors, but in community detection, nodes can share similar features. Second, we cannot take regularities and information for homogeneous segments from the image using just a single pixel. Third, compared with communities, images share some information, as an example, two adjacent regions belong probably to the same community. So, to address the mentioned problems, the proposed framework takes advantage of the inherent properties of the image and also the efficient optimization in modularity/stability using community detection algorithms. In this section, we describe the proposed framework steps so as the reader will have a global picture of the entire framework before delving into the details. We refer to Fig.1 for the illustration of the steps of the proposed general framework. The details of each step and some technical points are explained in the next sections.

\begin{figure*}[t!]
\begin{center}
\includegraphics[width=16cm,height=8cm]{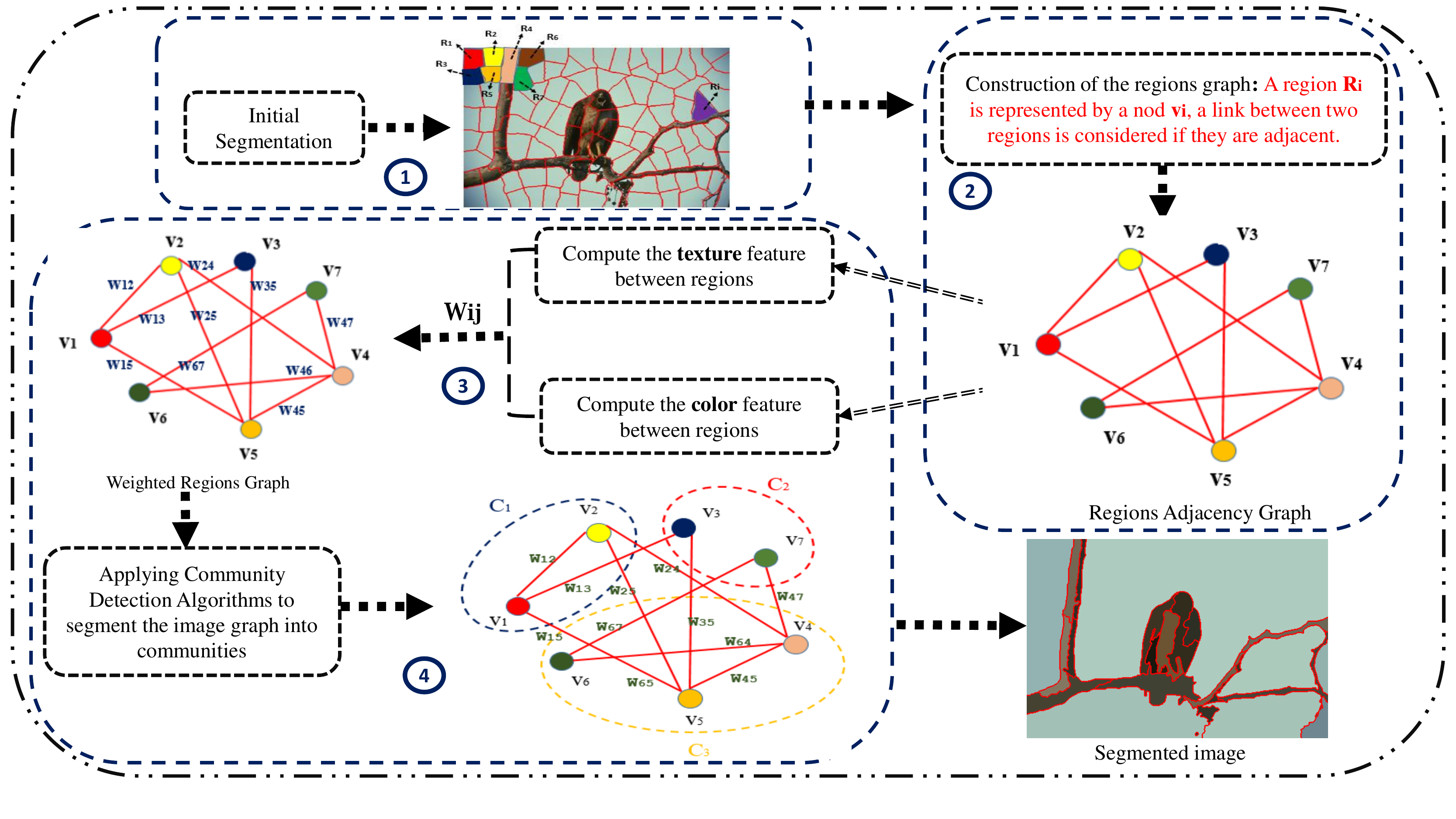}
\caption{ Flow chart of the proposed framework for an iteration.}
\end{center}
\end{figure*}

%
%
%
%

\subsection{Initial segmentation}

The initial segmentation process is very important to the function of the proposed framework. Because a single pixel cannot capture any information about texture, using regions instead of pixels capture the details among pixels (e.g texture) and prevent local variations to be lost. It showed that regions have the advantage to adapt themselves to the image structure, being larger where the color remains similar over a large area. Moreover, regions decrease the number of nodes in a graph abruptly from millions to thousands, hence, the computational complexity, without affecting the segmentation performance. For all these reasons we use an initial segmentation approach to provide very small regions of pixels that contain information and regularities and which will be used to compute the similarity between regions. In addition, the proposed framework allows using various methods that have been adopted to produce this over-segmentation.



\subsection{Regions Adjacency Graph construction}

Unlike conventional networks, images contain spatial a priori information compared to social networks or citation networks. Adjacent regions in the image are often considered as a single image segment, than other regions which are far away. So, we construct the RAG using the spatial a priori information of the image. Let $G = (V, E)$ be an undirected graph, where $v_i\in V$ is a
set of nodes corresponding to image regions $R_i$. E is a set of edges connecting the pairs of neighboring nodes. In other words, an edge is considered between two nodes, if their corresponding regions are adjacent in the image. As shown in Fig 3 the RAG is built after the initial segmentation, where each region in the image is considered as a node in the network.

\begin{figure*}[!h]
\begin{center}
\includegraphics[width=13cm,height=7cm]{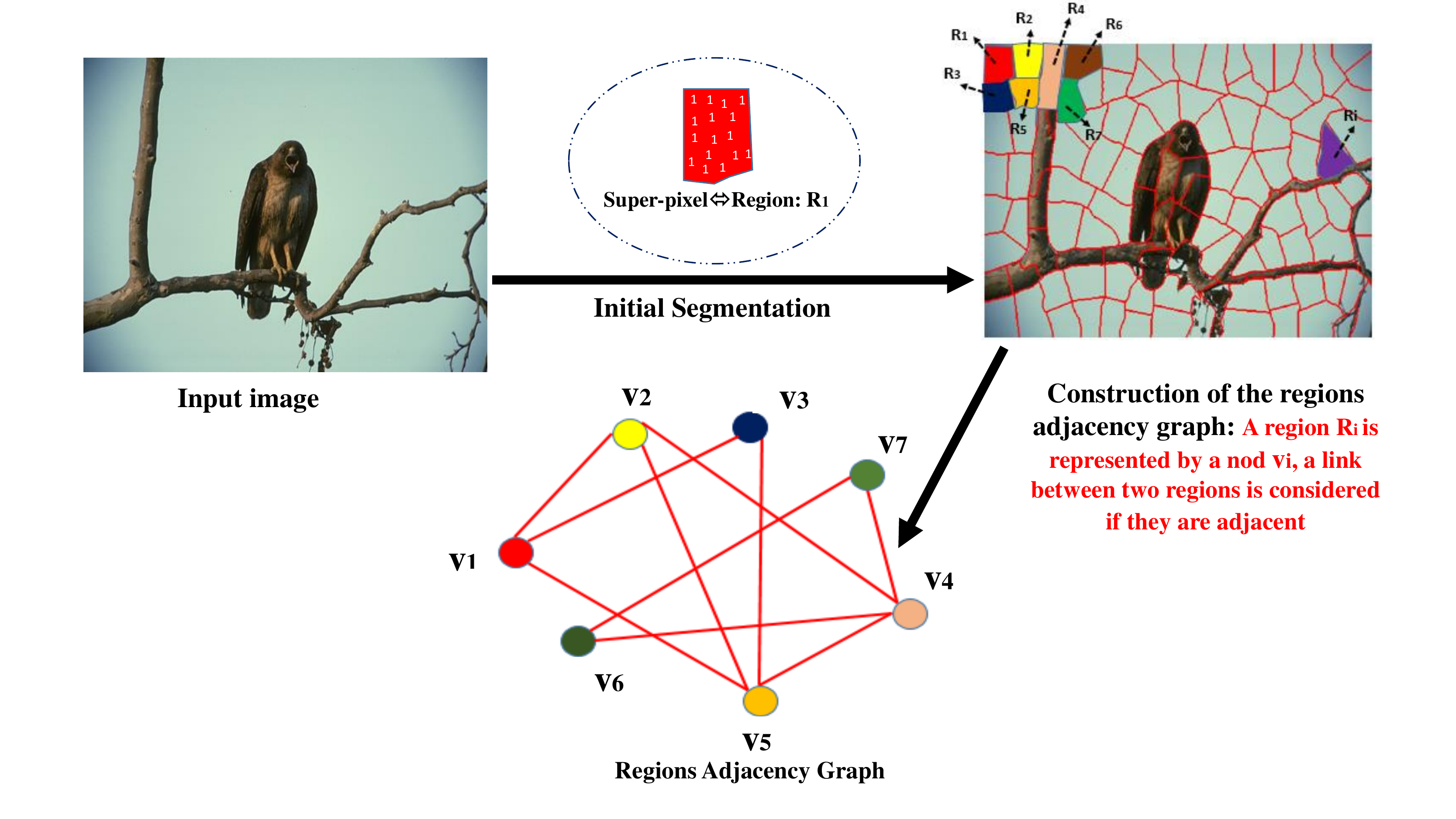}
\caption{ The construction process of the RAG from initially segmented images, each region $R_i$ in the image represents a node in the RAG.}
\end{center}
\end{figure*}

\subsection{Weighting the RAG}


%
 
 In this work, we compute the weight using the similarity between regions, where the color and texture feature of the image are employed between two adjacent regions in the RAG. To compute the similarity matrix W (weighted RAG), we use equation (2) and (4), for measuring the similarity between each two adjacent regions. Then, we associate a weight between them. Unlike the proposed approach in \cite{10} where authors built a graph by considering each node as a connected super-pixel according to a weight function, based only on color. In this work, the similarity is computed using a combination of the LAB and HOG features. We refer to \cite{15} RAG). In \cite{15}, authors use a hybrid model that combine two features. In this work we choose the texture (HOG) andfor the construction of the similarity matrix (weighted  the color (LAB) features as defined in the equation below:

\begin{equation}
\begin{split} 
{\rm W=w_{ij}= a\times \sqrt{t_{ij}\times c_{ij}}+(1-a)\times c_{ij}; }&\textbf{        }\\
{\rm(i,j)=1,..,n }& \textbf{        }
\end{split}
\end{equation}

 Where n denotes regions number and \textbf{$a$ }is a balancing parameter.







\subsection{Extracting communities from the network}

In this step, from the weighted RAG, we extract communities using community detection algorithms. To find the best partition of the network that gives a maximum modularity or stability, several algorithms have been proposed in the literature. Unlike the proposed approaches in Li \textit{et al.} \cite{6} and Abin \textit{et al.} \cite{7} which use community detection algorithms that have a high computational cost, and does not always produces the best segmentation, such as Newman-Fast algorithm and Modularity optimization. The proposed framework uses efficient community detection algorithms \cite{20,21,22,23} which strike the best balance between the computational cost and segmentation performance. Moreover, the proposed framework allows using any future existing community detection algorithms in this step.

\begin{algorithm}[h!]
\DontPrintSemicolon
\SetAlgoLined
\SetKwInOut{Input}{Input}\SetKwInOut{Output}{Output}
\Input{A color image \textit{I} }
\Output{The set of image segments $C_i = \{C_{i1}, ..., C_{ic}\}$ with $c \leq n$}

\BlankLine
Compute the initial set of region $R = \{R_1, ..., R_n\}$ where n is the number of regions \\
Initialize l=0\\
initialize the community structure $C_0 = \{C_{01}, ..., C_{0n}\}$ each region is assigned to a community\\
\Do{community structure still change ($C_{l} \neq C_{l-1}$)}
{    
    
    \BlankLine
    Construct the RAG; \\
    Assign a node to each region
    Assign an edge between two nodes if they are adjacent\\
    Compute the texture and the color feature for each region $R_i$;\\
    Compute the weight of the RAG (W) according to Equation (10), $w_{ij}\neq 0$ only if $R_i$ and $R_j$ are 
    adjacent regions in RAG; \\
     Compute the community structure of the weighted RAG using a community detection algorithm $C_l = \{C_{l1}, ..., C_{lm}\}$ where m is the   current number of communities;\\
     Merge the regions that belongs to the same communities;\\
     l=l+1;

}

\caption{  }
\end{algorithm}

\subsection{Merging process}

In this step, an iterative process is used to construct the similarity matrix, and to recalculate weights between regions at each iteration according to the equation (10). Because when we use community detection algorithms, regions keep expanding at each iteration, and the similarity measure given by the previous iteration may be is not suitable for the current iteration. So, updating the weighted RAG at each iteration allows reevaluating weights between regions. This process avoids many small regions in the image which should be merged on the same community according to the perspective of the human visual system.

The algorithm of the proposed framework can be summarized in Algorithm.1.


\begin{figure*}[t!]
\begin{center}
\begin{tabular}{c}
   \includegraphics[width=13.5cm,height=4.5cm]{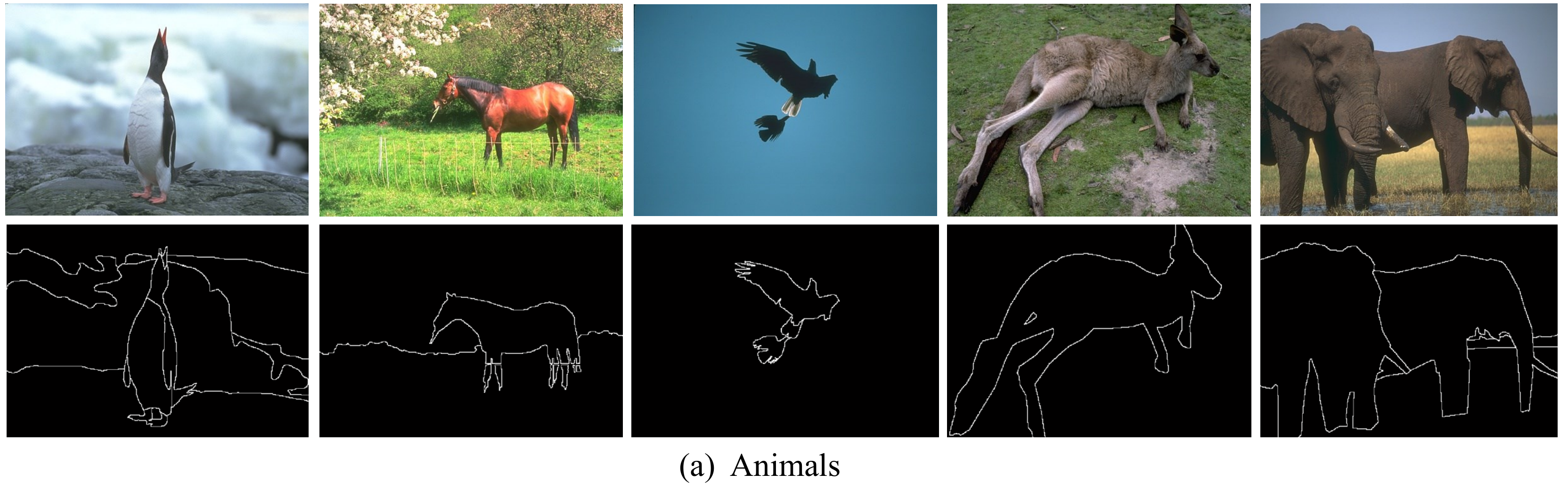} \\
   \includegraphics[width=13.5cm,height=4.5cm]{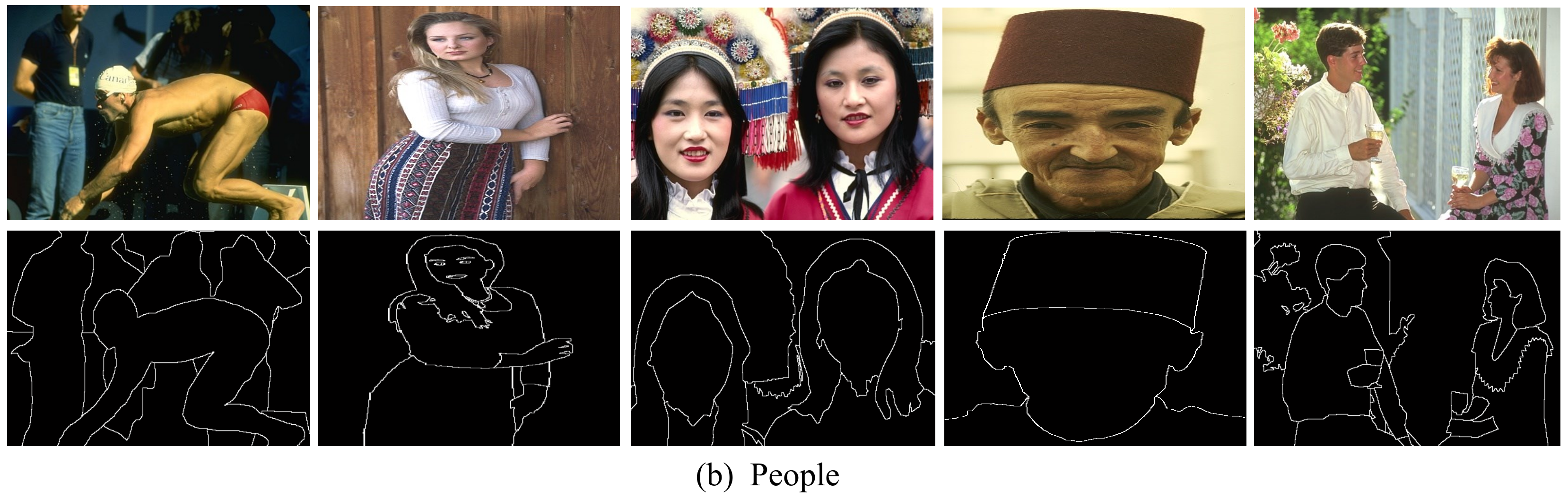} \\
   
    \includegraphics[width=13.5cm,height=4.5cm]{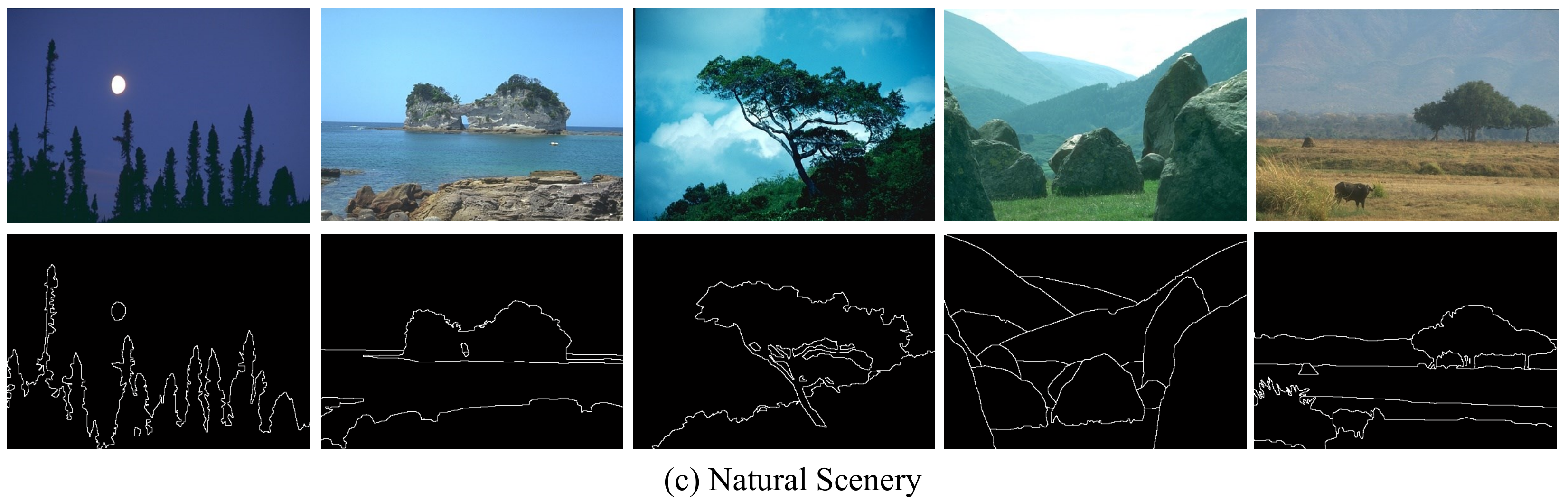} \\
   \includegraphics[width=13.5cm,height=4.5cm]{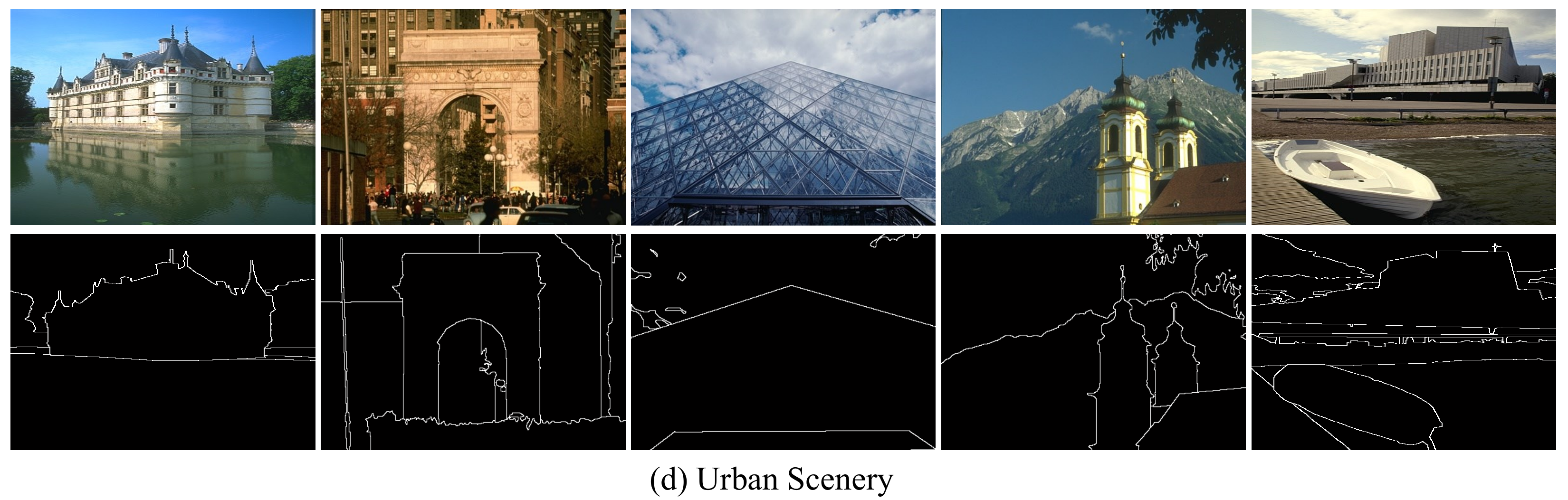}  \\

\end{tabular}
  
   \caption{\textbf{BSDS500 images for different categories.} For each category, Line 1: Original images. Line 2: Ground truths segmentation.}
\end{center}
\end{figure*}

\section{Data and Methods}

\subsection{Data}

In order to check the performance of the proposed framework in comparison with other alternative methods, manual segmentation labeling of a database is required. Thanks to its availability, the publicly available Berkeley Segmentation Data Set 500 (BSDS500) \cite{25} was used to evaluate the performance of the proposed framework, BSDS500 contains 500 natural images, including 200 images for training, 200 images for testing and 100 images for validation. Boundaries are labeled for each of the 500 images of size 481$\times$321 by several workers and are averaged to form the ground truth. Figure 3 shows BSDS500 images for different categories, with their ground truths segmentation. As for the evaluation of segmentation performance, we use three different metrics, i.e., the  Probability Rand Index (PRI), Variation of Information (VOI), Precision, Recall and F-measure. These metrics have different characters, for example, the PRI tends to over-segmentation, while VOI encourage under-segmentation. Therefore, an overall consideration of these three metrics is necessary and reasonable. Note that, good segmentation corresponds to high PRI, Precision, Recall and F-measure value, but low VOI value. All algorithms are implemented in Matlab and are carried out on 4 GB of RAM and a 2.60 GHz processor.

\subsection{Methods}
To investigate the efficiency of the proposed framework, we study first the influence of some parameters used computing the similarity between regions. We perform first experiments on 100 images of BSDS500 using empirically several values of the balancing parameter a to find out the appropriate value that achieve the best segmentation results. 

Second, a comparison between image features (Color and Texture) is done. In this regard, we examine the performance of the framework quantitatively as well qualitatively, in cases where texture alone, color alone and color-texture are used in the segmentation process. 

Third, in order to choose the appropriate initial segmentation algorithm for the framework, we run some experiments in 100 images of BSDS500 and we compare the average values of the PRI, VOI, Precision, and Recall for each algorithm. Then, we choose the algorithm which ensure the best segmentation results. 

We study next, the influence of community detection algorithms. We consider that all images in BSDS500 are classified into four categories, namely, people, urban scenery, animals and natural scenery. We take for each category five randomly chosen images with their ground truths, and we compare the results of four efficient algorithms qualitatively as well quantitatively using the average value of PRI, VOI, Precision and Recall metrics for each category. 

Finally, with the appropriate parameters and methods discussed previously, we perform a qualitative and quantitative comparison of the framework with alternative methods: Li \textit{et al.} \cite{6}, Abin \textit{et al.} \cite{7}, Lossy Compression (LC) \cite{28} and EDISON \cite{29}. In Li \textit{et al.} \cite{6} and  Abin \textit{et al.} \cite{7}, we preserve the same parameters used by authors. In EDISON \cite{29} method which is based on the mean shift implementation in both boundaries extraction and noise filtering scheme, the main parameters of EDISON is the minimal region size. So, we set the parameter value to 1000, to avoid the creation of small regions. In Lossy Compression (LC) \cite{28} we use the Gaussian Mixture Model to fit the image textures, and for finding the optimal segmentation we employ the principle of Minimum Description Length, that produces the minimum coding length under a certain distortion ratio. We use the distortion rate $\epsilon$= 0.2.

\subsection{Evaluation metrics}

For our evaluation, we investigate for the quantitative evaluation, the Probabilistic Rand Index (PRI) \cite{26} and the Variation of Information (VOI)\cite{27} which are a well-known evaluation metrics for segmentation. The PRI measures the probability that a segmentation and its ground truth have matched labels in the two partitions. The larger the value is, the more the similarity between the two segmentation is. The PRI range is in [0,1]. 

VOI metric measures the sum of information gain and information loss between two segmentations. The VOI metric is nonnegative, the more is lower the more the similarity is greater. It's defined by the formula below:

\begin{equation}
{\rm VOI(C,C') = H(C) + H(C')\textbf{    -   }  2I(C,C')}
\end{equation}
where H(C) and H(C') denotes the entropy of the two segmentation C and C' respectively and I(C,C') denote the mutual information of C and C'. The metric range is $[0,\infty]$, and the smaller the value is, the more similar the two segmentations are.\\

We also evaluate the performance of the proposed framework from two aspects: Precision and Recall. These two measures are attractive as measures of segmentation quality because they are sensitive to under and over-segmentation, under-segmentation leads to low recall scores, while over-segmentation leads to low precision scores. \\ \\
The Precision measures the fraction of detected boundary pixels which match the ground-truth boundaries is defined as: 

\begin{equation}
{\rm Precision= \frac{|S_{test}|\cap |S_{gt}|}{|S_{test}|}}
\end{equation}

where $S_{gt}$ is the ground truth segmentation and $S_{test}$ the testing segmentation and $|S|$ denotes the boundary pixels number in the segmentation S.

The Recall computes the percentage of ground-truth boundary pixels that are detected, is defined as:

\begin{equation}
{\rm Recall= \frac{|S_{test}|\cap |S_{gt}|}{|S_{gt}|}}
\end{equation}

$F_{\alpha}$-measure  is a quality measure based on Recall and Precision only, which measure the harmonic mean of the Precision and Recall, is defined as:

\begin{equation}
{\rm  \textit{\rm F-measure}= \frac{Precision . Recall}{(1-\alpha). Recall+ \alpha . Precision}}
\end{equation}

For all next experiments we set $\alpha=0.5$.

\section{Experimental results}

We have discussed the proposed framework but so far not shown any results. For the sake of completeness and illustration, in this section, the performance of the proposed framework is assessed qualitatively as well as quantitatively by providing some experiments.

\subsection{Influence of similarity measures}

\subsubsection{Adjusting the balancing parameter a}
As mentioned in section 5.3, during each iteration, we use a combination of the LAB feature and the HOG texture feature to compute the similarity between regions, using the equation (10), where a is a balancing parameter. If $a = 0$, it means that the texture information is not considered. We can observe in figure 4 that the wheel of the vehicle and its upper are well encoded into the similarity. With increasing the value of a, more such information patterns are encoded into the similarity, thus better preserves the regularities. However, if a is too large in our case a=0.8, the wheel and the upper of the vehicle in the image are merged into one segment. We run the framework to determine the best value of the parameter a, we vary a from 0 to 1 with 0.2 intervals because when we step by 0.1 or 0.05 we can't observe any change in the segmentation result. Results show that a = 0.4 gives the best performance in term of both metrics PRI and VOI. In all next experiments, we use a=0.4.

\begin{figure*}[h!]
\begin{center}

\includegraphics[width=12cm,height=3cm]{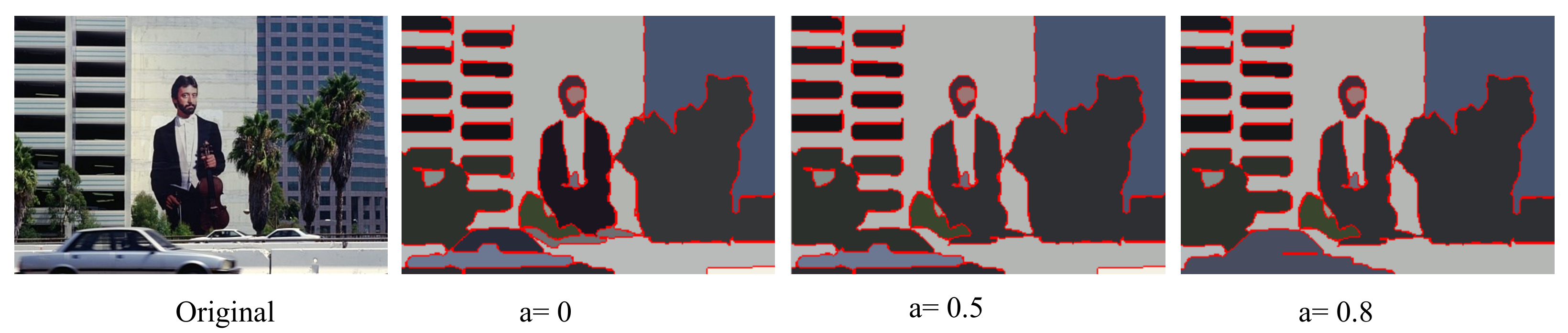}
\caption{Image segmentation with various values of the balancing parameter a by FMCDRN.}
\end{center}
\end{figure*}

\subsubsection{Comparison between features of similarity}

An important issue for the proposed framework is to evaluate the influence of the color(LAB) and texture(HOG) information in the segmentation process. We compare the performance of the proposed framework in cases where HOG alone, LAB alone and the combination HOG+LAB are used in the segmentation process. Our experiments was tested in 100 images of Berkeley segmentation dataset using meanshift as an initial segmentation and FMCDRN as a community detection algorithm. The balance between the texture and color is performed by the weight $w_{ij}$ in equation (10) and to obtain the texture and color alone segmentation we override the balancing parameter a with manual settings (i.e. a=0 for color alone segmentation and a=1 for texture alone segmentation). Since in BSDS500, there are multiple segmentation maps of the ground-truth, 5 segmentation maps for each image, in our experiment the mean value of the computed metrics is used between all the segmentation maps for each image and the segmentation result. As illustrated in Table 1 which presents the average values of the PRI, VOI, Precision and Recall for 100 images, we can notice that texture and color alone results are generally inferior to results obtained from the combination texture and color for all metrics.

\begin{table}[h!]
\caption{Quantitative comparison between texture and color features.}
\begin{tabular}{l|l|l|l|l|l|}
\cline{2-6}
 & PRI & VOI & Precision & Recall & F-measure \\ \hline
\multicolumn{1}{|l|}{Framework with Texture (HOG)} & 0.708 & 1.808 & 0.698 & 0.571 & 0.628 \\ \hline
\multicolumn{1}{|l|}{Framework with Color(LAB)} & 0.715 & 1.794 & 0.685 & 0.589 & 0.633 \\ \hline
\multicolumn{1}{|l|}{Framework with Texture-Color(HOG+LAB)} & 0.828 & 1.695 & 0.788 & 0.621 & 0.694 \\ \hline
\end{tabular}
\end{table}

Figure 5 shows sample visual results for the segmentation of three images of BSDS500. We can observe that using only one of the proposed features leads to severe performance degradation, for example in the first image it is clear that with color or texture only used in computing the similarity, the visually coherent pyramid together with the desert are broken, the same thing for human face. Because even with properly chosen distance parameter, using only one of the feature, color or texture, breaks the regularities inside the object and leads to over-segmentation. In contrast, the combination of these two features leads to better visual performance and well preserves these regularities.

\begin{figure*}[ht!]
\begin{center}

\includegraphics[width=12cm,height=6cm]{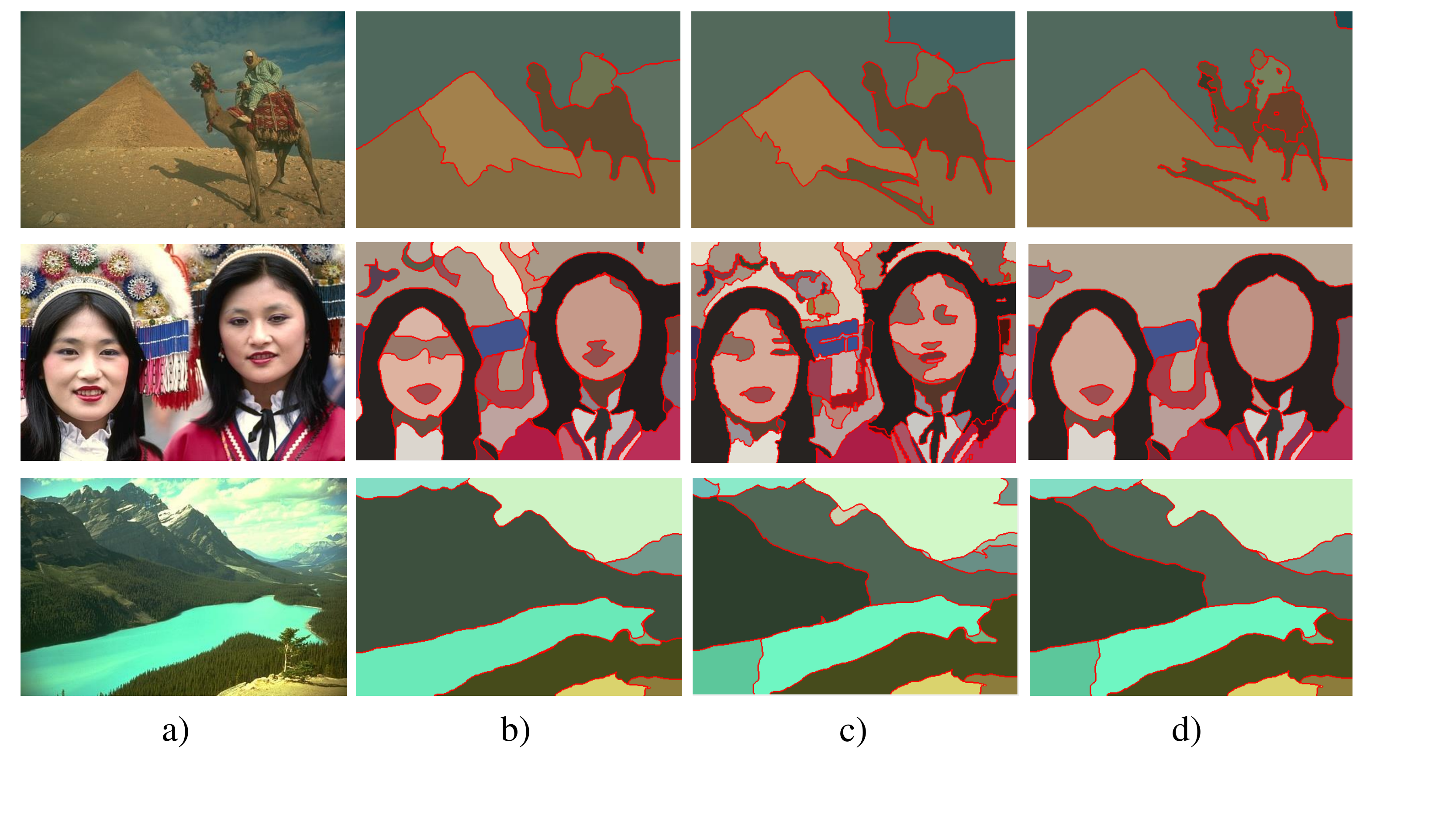}
\caption{Comparison of segmentation results: a) Original image; b) HOG feature; d) LAB feature; c) HOG with LAB. }
\end{center}
\end{figure*}

\subsection{Influence of the Initial Segmentation Algorithms}

To choose the appropriate initial segmentation that ensure the best segmentation results, experiments have been conducted using using two initial segmentation algorithm to find out the appropriate one. The parameters for the Superpixels algorithm were chosen to give a number of regions in the same range as those for the Meanshift algorithm. For each image we compute the value of PRI, VOI, Precision, and Recall, then we average all these metrics for 100 images of BSDS500. Table 2 reports the average values of the PRI, VOI, Precision, and Recall of all images for Meanshift and Superpixels algorithms. We can notice that the Meanshift approach is more appropriate than the super-pixels for all the four metrics. Motivated by the limitations exposed in the cited approaches for segmentation, we want that the proposed framework takes the time complexity aspect into consideration. To achieve that, we compare the run time of each algorithm. Results show that Meanshift runs faster than superpixels, about 2.5 faster than superpixels. For all these reasons, Meanshift is chosen as an initial segmentation for the next experiments.

\begin{table}[h!]
\caption{Quantitative comparison between Meanshift and Superpixels.}

\begin{tabular}{l|l|l|l|l|l|l|}
\cline{2-7}
                                                 & PRI   & VOI   & Precision & Recall & F-measure & Running time (s)\\ \hline
\multicolumn{1}{|l|}{Framework with Meanshift}   & 0.828 & 1.695 & 0.788     & 0.621  & 0.694     & 4.465                  \\ \hline
\multicolumn{1}{|l|}{Framework with Superpixels} & 0.791 & 1.924 & 0.731     & 0.574  & 0.642     & 10.531                 \\ \hline
\end{tabular}
\end{table}

\subsection{Influence of community detection algorithms}

In this section, we compare the proposed community detection algorithms, to choose the best of them for the next comparison with alternative methods. 

In the first qualitative evaluations experiment, as shown in Fig.6, 7, 8 and 9, for each category Animals, People, Natural Scenery and Urban Scenery, we test the proposed framework for each community detection algorithms. The results produce sizable regions and give much better results for all selected image for each category, for example, the human face in the people category, animals in natural scenery category, the castle in urban scenery category and mountains in natural scenery category. As observed in figures the Infomap algorithm doesn't always gives the best segmentation and underestimates the number of communities even though is faster than other community detection algorithms. In addition, we can observe from the figures that the FMCDRN algorithm gives the best segmentation of the image.

We also assess the performance of the proposed framework with the four segmentation techniques quantitatively. Tables 3, 4, 5 and 6 present the average values of the PRI, VOI, Precision and Recall for each category. Again, the results of the FMCDRN algorithm show their efficiency for the image segmentation task in term of PRI/VOI/Precision/Recall/F-measure.

\begin{table*}[h!]
\caption{Quantitative comparison between community detection algorithms used in the proposed framework on \textbf{animals} category.}
{\begin{tabular*}{15pc}{@{\extracolsep{\fill}}|l|C{2cm}|C{2cm}|c|c|c|C{2cm}|}
  \hhline{-------}

Algorithms & PRI (larger better)& VOI (smaller better)& Precision &  Recall &  F-measure & Running Time (second)\\
 
   \hhline{-------}

 Framework+FMCDRN &\textbf{0.911}&  \textbf{1.520}& \textbf{0.897}   & \textbf{0.789}&     \textbf{0.839}&  4.324\\
  Framework+FGMDO &  0.858& 1.585 &  0.868 & 0.681& 0.763& 8.583\\
  Framework+Louvain &  0.847& 1.610& 0.857  & 0.670& 0.752& 5.322\\

  Framework+Infomap  &  0.588& 2.402&  0.598    & 0.546& 0.570 & 2.161\\
 
   \hhline{-------}

\end{tabular*}}{}
\end{table*}

\begin{table*}[!t]
\caption{Quantitative comparison between community detection algorithms used in the proposed framework on \textbf{people} category.}
{\begin{tabular*}{15pc}{@{\extracolsep{\fill}}|l|C{2cm}|C{2cm}|c|c|c|C{2cm}|}
  \hhline{-------}
Algorithms & PRI (larger better)&  VOI (smaller better) & Precision& Recall&  F-measure& Running Time (second)\\
\hhline{-------}

 Framework+FMCDRN &  \textbf{0.921}&  \textbf{1.510}&  \textbf{0.901}   & \textbf{0.887}&  \textbf{0.893} &  4.525\\
 Framework+FGMDO & 0.868&  1.543 &0.878 & 0.696&    0.776&8.842\\
  Framework+Louvain &  0.849&   1.598 & 0.864   & 0.684& 0.763 &5.762\\
 
 Framework+Infomap  &0.562&  2.454& 0.558  & 0.539&  0.548 & 2.258\\
  \hhline{-------}

\end{tabular*}}{}
\end{table*}

\begin{table*}[!t]
\caption{Quantitative comparison between community detection algorithms used in the proposed framework on \textbf{natural scenery} category.}
{\begin{tabular*}{15pc}{@{\extracolsep{\fill}}|l|C{2cm}|C{2cm}|c|c|c|C{2cm}|}
  \hhline{-------}

 Algorithms &  PRI (larger better)&  VOI (smaller better) & Precision &  Recall&  F-measure & Running Time (second)\\   \hhline{-------}

  Framework+FMCDRN &  \textbf{0.887}&  \textbf{1.610}& \textbf{0.891}  & \textbf{0.837}&     \textbf{0.863} & 4.897\\
  Framework+FGMDO &  0.849&   1.643 &  0.862  & 0.674&    0.756& 8.984\\
  Framework+Louvain & 0.828&   1.648 & 0.856   &  0.671&  0.754& 5.954\\

  Framework+Infomap  &  0.577&  2.421 & 0.579    & 0.564&   0.571&2.742\\
  \hhline{-------}

\end{tabular*}}{}
\end{table*}

\begin{table*}[!t]
\caption{Quantitative comparison between community detection algorithms used in the proposed framework on \textbf{urban scenery} category.}
{\begin{tabular*}{15pc}{@{\extracolsep{\fill}}|l|C{2cm}|C{2cm}|c|c|c|C{2cm}|}
  \hhline{-------}

 Algorithms &  PRI (larger better)&  VOI (smaller better) & Precision &  Recall &  F-measure & Running Time (second)\\    \hhline{-------}

 Framework+ FMCDRN & \textbf{0.887}& \textbf{1.610}&  \textbf{0.891}  &  \textbf{0.837} &    \textbf{0.863}& 4.624\\
  Framework+FGMDO &  0.789&   1.697&  0.782& 0.621&  0.692& 8.725\\
  Framework+Louvain &  0.783&   1.698& 0.779  &  0.612 &  0.685& 5.689\\

  Framework+Infomap & 0.574&  2.419 &  0.585  & 0.556&  0.570& 2.521\\

  \hhline{-------}

\end{tabular*}}{}
\end{table*}

\begin{figure*}[h!]
\begin{center}
\includegraphics[width=13cm,height=10cm]{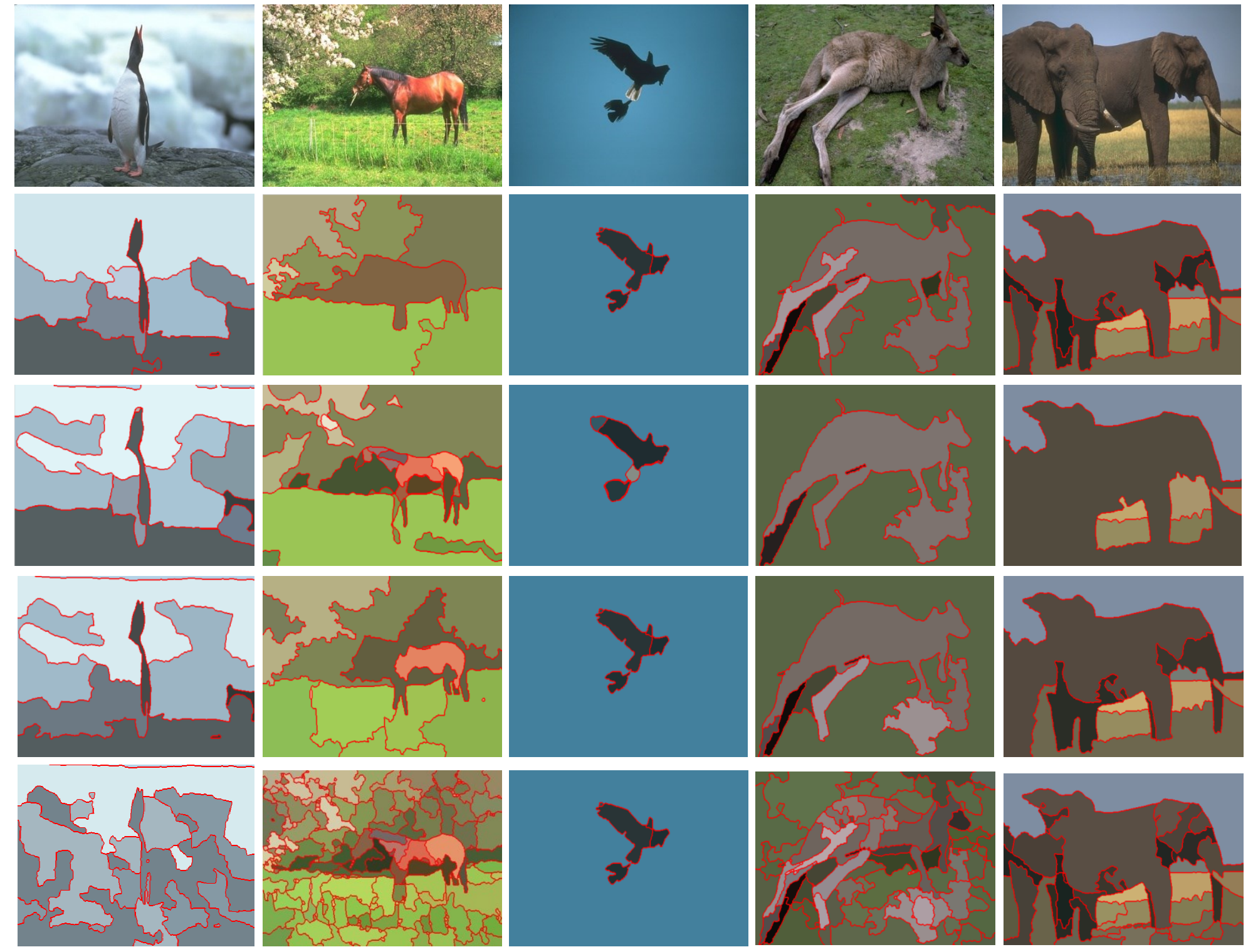}
\caption{Segmentation results of the framework with the proposed community detection algorithms, for images from \textbf{animals} category, Line 1:Original image; Line 2: FMCDRN algorithm; Line 3: FGMDO algorithm; Line 4: Louvain; Line 5: Infomap.}
\end{center}
\end{figure*}

\begin{figure*}[h!]
\begin{center}
\includegraphics[width=13cm,height=10cm]{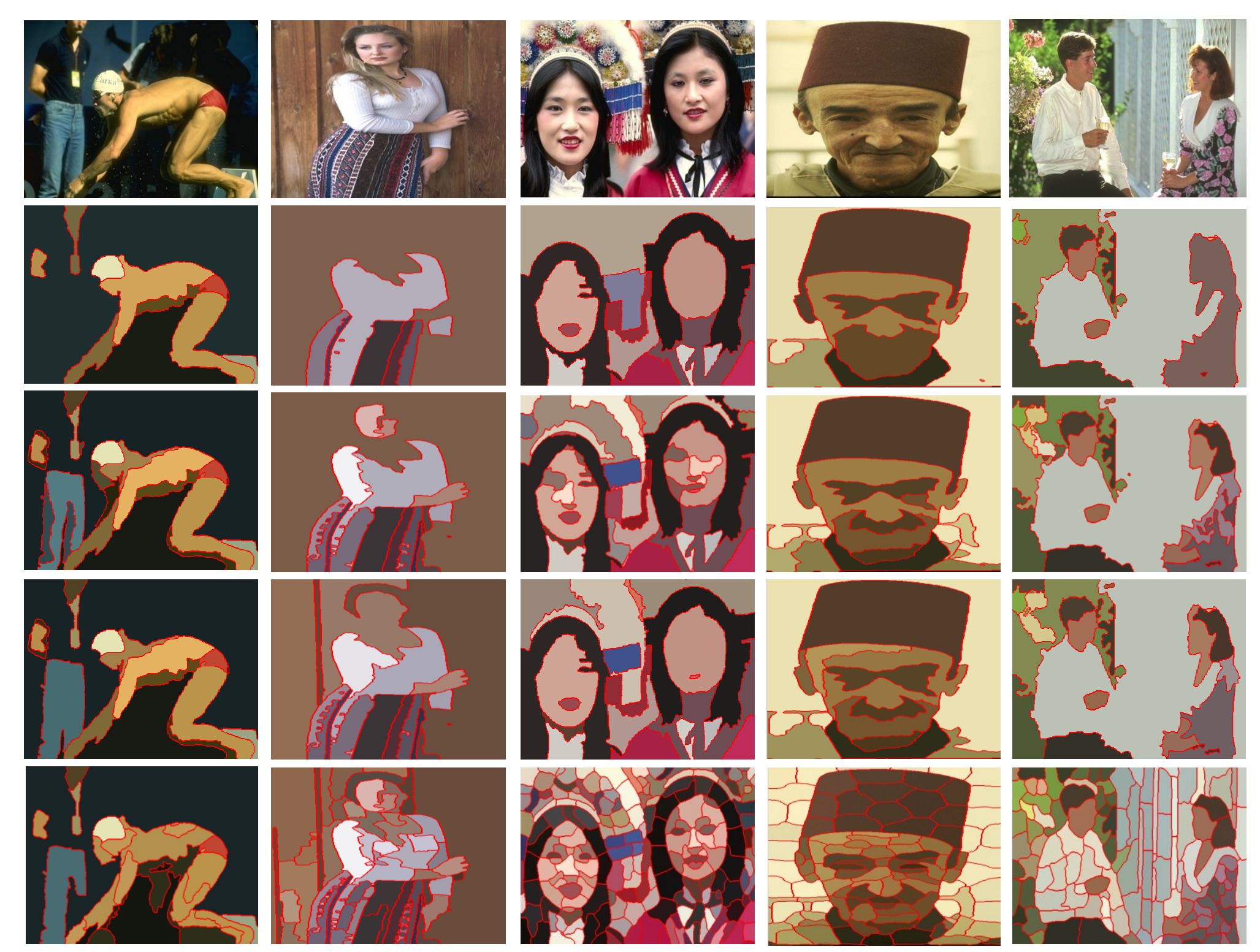}
\caption{Segmentation results of the framework with the proposed community detection algorithms, for images from \textbf{people} category, Line 1:Original image; Line 2: FMCDRN algorithm; Line 3: FGMDO algorithm; Line 4: Louvain; Line 5: Infomap.}
\end{center}
\end{figure*}

\begin{figure*}[h!]
\begin{center}
\includegraphics[width=13cm,height=10cm]{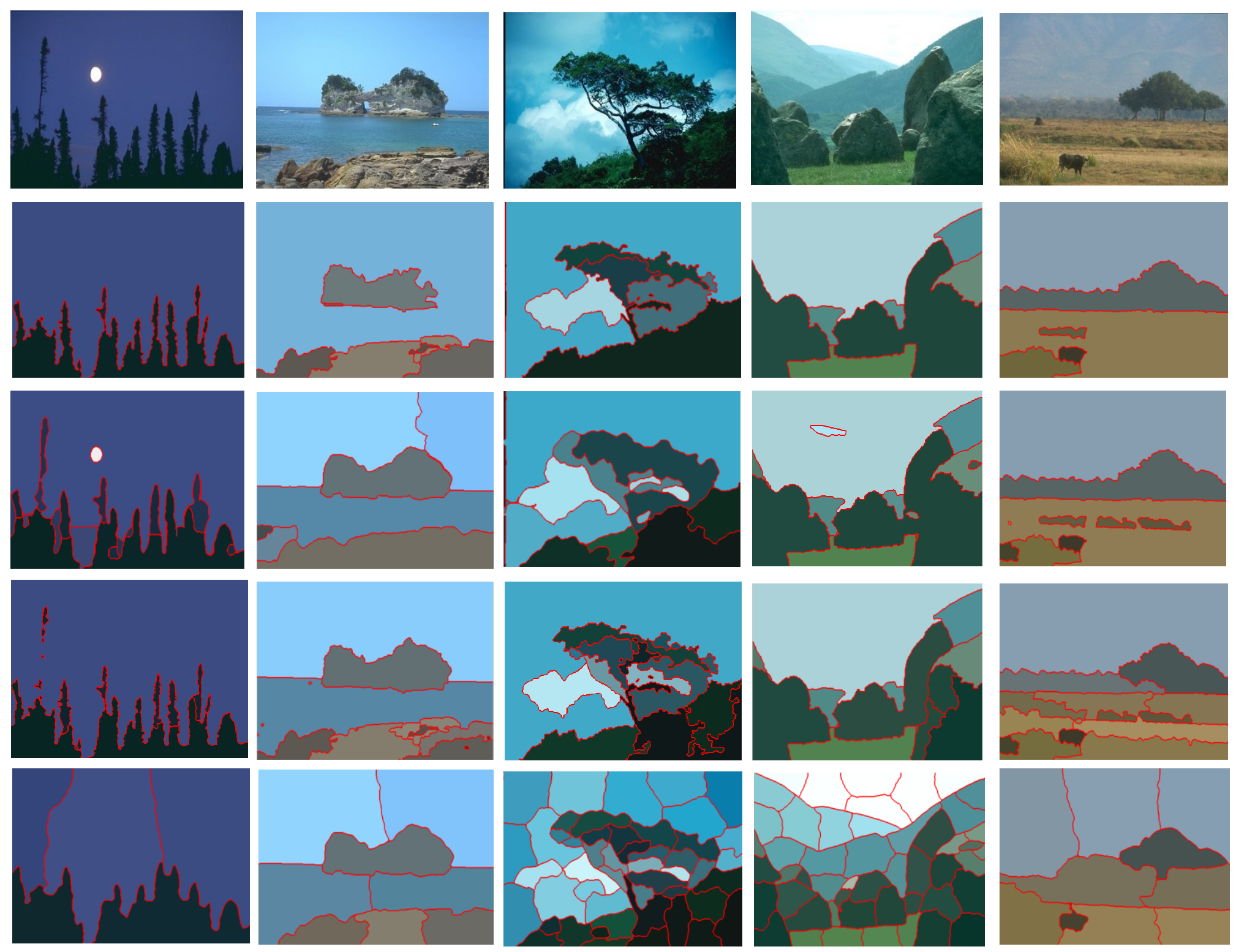}
\caption{Segmentation results of the framework with the proposed community detection algorithms, for images from \textbf{natural scenery} category, Line 1:Original image; Line 2: FMCDRN algorithm; Line 3: FGMDO algorithm; Line 4: Louvain; Line 5: Infomap.}
\end{center}
\end{figure*}

\begin{figure*}[h!]
\begin{center}
\includegraphics[width=13cm,height=10cm]{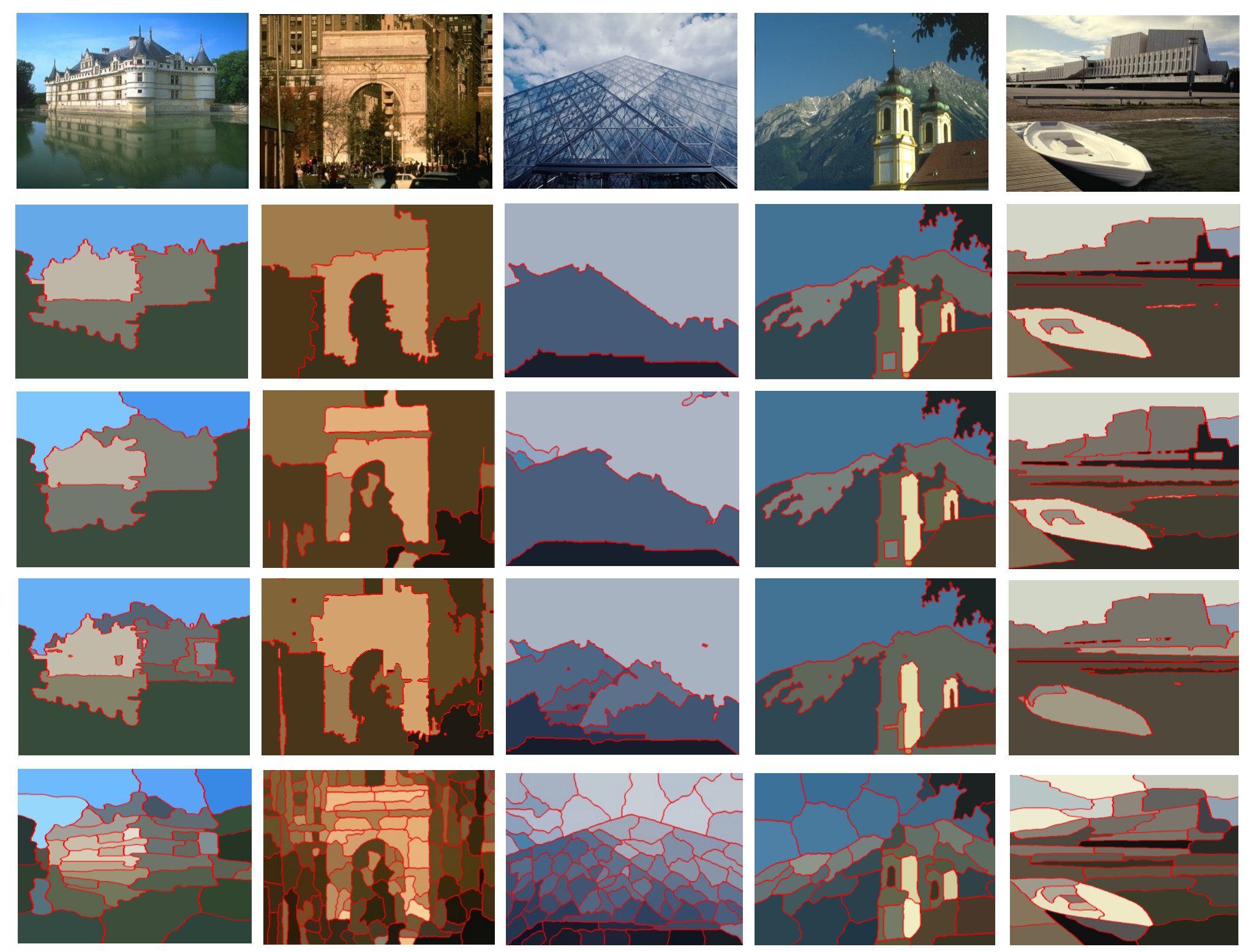}
\caption{Segmentation results of the framework with the proposed community detection algorithms, for images from \textbf{urban scenery} category, Line 1:Original image; Line 2: FMCDRN algorithm; Line 3: FGMDO algorithm; Line 4: Louvain; Line 5:Infomap.}
\end{center}
\end{figure*}

  

  
\subsection{ Comparison with the alternative methods}

First, we perform a qualitative comparison of the framework with some well-known state of the art segmentation methods: Li \textit{et al.} \cite{6}, Abin \textit{et al.} \cite{7}, Lossy Compression (LC) \cite{28} and EDISON \cite{29}. We choose FMCDRN algorithm for the proposed framework in comparison, because it gives the best image segmentation as shown previously. Figure 10 shows that LC, EDISON shows the different extent of over-segmentation by resulting many small regions, and also by breaking information and regularities in some homogeneous regions of the image, compared to the proposed framework, which preserves information and regularities in the segmented image and produces sizable homogeneous regions, also the framework has the best performance as Li \textit{et al.} \cite{6} and Abin \textit{et al.} \cite{7}. Results indicate the superiority of the proposed framework over other methods.
 We compute next the average values of the PRI, VOI, Precision and Recall for all images. As shown in table 7 the framework gives a high value and better results for the image segmentation task, compared to all well-known segmentation algorithms  EDISON, LC, Li \textit{et al.} \cite{6} and Abin \textit{et al.} \cite{7} in term of PRI/VOI, and also have a close performance to human visual perception with PRI=0.828 and VOI=1.695. Also, the Precision, Recall and F-measure of the proposed framework with FMCDRN algorithm obtain the highest values with Precision=0.788, Recall=0.621 and F-measure=0.694 compared to the other algorithms which indicate that most of our segmentation has consistent labels with the ground-truth segmentation in BSDS500. As a conclusion, we can say that the proposed framework achieves better performance in terms of Precision, Recall, and F-measure compared with other state of the art algorithms.

  \begin{figure*}[h!]
\begin{center}
\includegraphics[width=13cm,height=17cm]{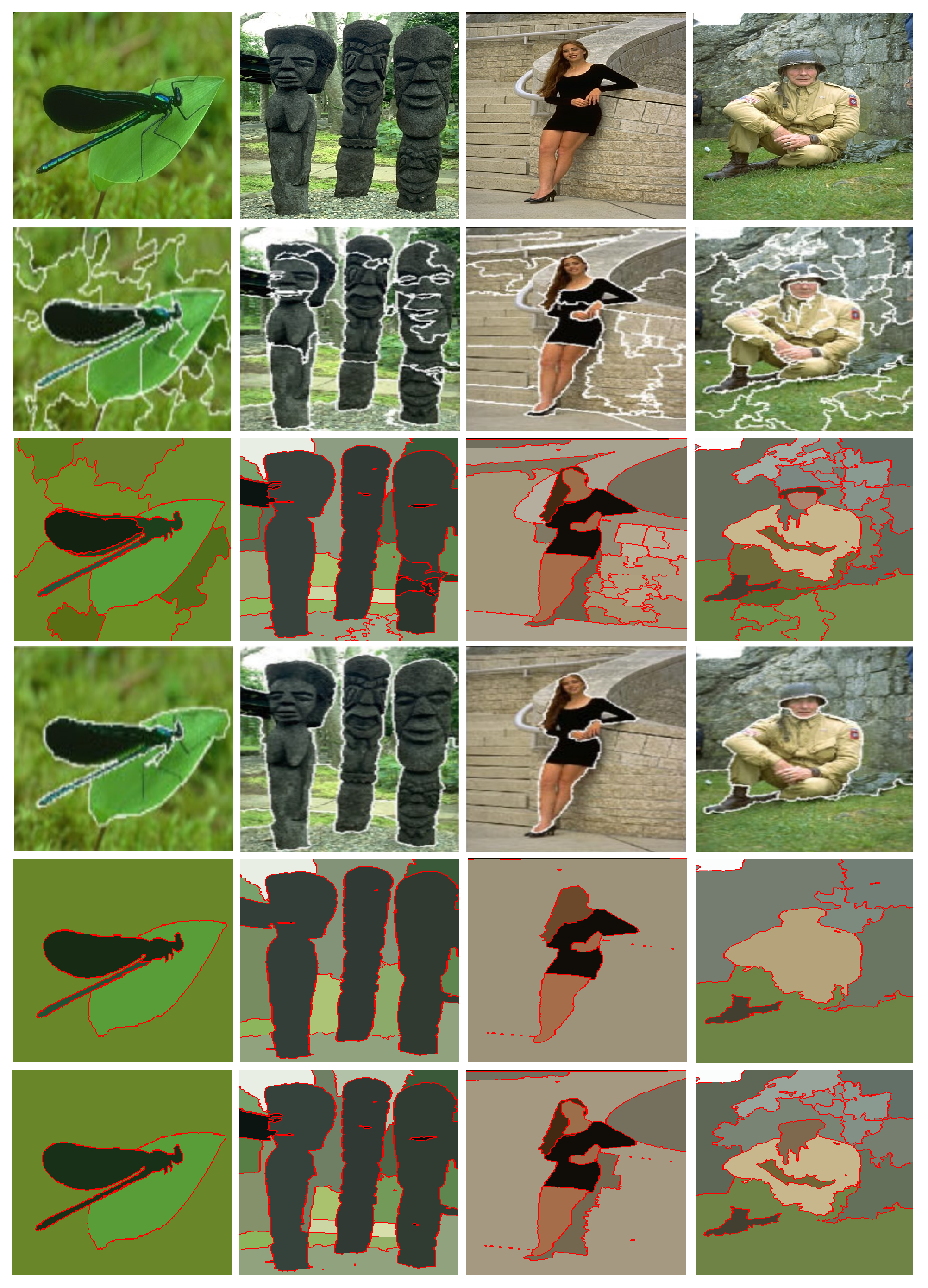}
\caption{Comparison of segmentation results of all algorithms, Line 1:Original image; Line 2:EDISON; Line 3: LC; Line 4: Abin \textit{et al.} \cite{7}; Line 5:Li \textit{et al.} \cite{6}; Line 6: Our Framework + FMCDRN algorithm.}
\end{center}
\end{figure*}

\begin{table*}[h!]
\caption{Quantitative comparison between different algorithms on all Berkeley dataset images.}
{\begin{tabular*}{15pc}{@{\extracolsep{\fill}}|l|c|c|c|c|c|}
  \hhline{------}

 Algorithms &  PRI (larger better)& VOI (smaller better) & Precision& Recall &  F-measure\\
  \hhline{------}

 Humain&  0.870&   1.160&  0.910 &  0.720 &0.797\\
 EDISON \cite{29} &  0.786& 2.002& 0.728& 0.524& 0.609\\
LC \cite{28}&  0.735&  1.978&  0.700& 0.531 &0.603\\
Abin \textit{et al.} \cite{7}&  0.813&  1.721&  0.764& 0.615 &0.681\\ 
Li \textit{et al.} \cite{6}&  0.777&  1.879&  0.733& 0.508 &0.600\\ 
 Framework+FMCDRN & \textbf{ 0.828}& \textbf{  1.695}& \textbf{0.788}& \textbf{0.621}&    \textbf{0.694}\\
  \hhline{------}

\end{tabular*}}{}
\end{table*}

\subsection{Computational Time}
We compare the running time between our proposed framework with FMCDRN algorithm and Li \textit{et al.} \cite{6} and Abin \textit{et al.} \cite{7}. Each algorithm is tested over 100 validation images of Berkeley dataset, and then for each step (Initial segmentation, Graph generation, and Community detection), we compute the mean running time. It can be observed in table 7 that the proposed framework runs consistently faster than Li \textit{et al.} \cite{6} and Abin \textit{et al.} \cite{7}, specifically, about 2.5 times faster than Abin \textit{et al.} \cite{7}, and 5 times faster than Li \textit{et al.} \cite{6}. As a conclusion, we can say that the proposed framework yields better results with shorter processing time than other algorithms of the stat of the art.

\begin{table*} [h!]
\caption{COMPUTATIONAL TIME OBTAINED IN THE SEGMENTATION OF ALGORITHMS (IN UNIT OF SECONDS).}
{\begin{tabular*}{15pc}{@{\extracolsep{\fill}}l|c|c|c|c|c}

  \hhline{------}

  \multirow{2}{*}{Algorithms} & \multicolumn{5}{c}{Computational Times} \\ \\
  \hhline{------}
  
 &\multicolumn{1}{c|}{Superpixels}  & \multicolumn{1}{c|}{MeanShift }&\multicolumn{1}{c|}{ Graph} &\multicolumn{1}{c|}{ Community Detection Algorithm} &\multicolumn{1}{c}{ Total}\\
 \hhline{~-----} 
Abin \textit{et al.} \cite{7}&  -&  0.239&  0.128&11.215 &11.482\\ 
Li \textit{et al.} \cite{6}& 3.299&  -&  0.231& 16,870  &20.40\\ 
 Framework+FMCDRN &  -&  0.239& 0.178& 4.026& \textbf{4.443}\\

  \hhline{------}

\end{tabular*}}{}
\end{table*}

\section{Conclusion}

This paper proposed a framework for image segmentation which takes advantages of the inherent properties of images and the optimization of modularity/stability. Efficient community detection algorithms are used to optimize modularity/stability, as FMCDRN, FGMDO, and Louvain. All these algorithms can detect automatically the number of the region in the image. By using both, Histogram of Oriented Gradients (HOG) texture feature and color feature, the similarity matrix is constructed adaptively between different regions by optimizing the modularity/stability and merge adjacent regions iteratively. If no change occurs in community structure when we apply community detection algorithms, the optimal segmentation is achieved. Our experiments have shown that the proposed framework gives a best qualitative segmentation result, as proved in the figures and achieve the best performance quantitatively compared to all state of the art methods in terms of PRI, VOI, Precision, and Recall. Since, the general framework based on three efficient community detection algorithms, it avoids the problem of having many small regions in the image and preserves information and regularities in the object. In addition, it provides a good time complexity and runs consistently faster than the state of the art algorithms.


\begin{thebibliography}{1}


\bigskip


\bibitem{1} Chen, Xiaohui, et al. "Image segmentation using a unified Markov random field model." IET Image Processing (2017).

\bibitem{2} J. A. Noble and D. Boukerroui, "Ultrasound image segmentation: A survey,"  IEEE Transactions on medical imaging, 25(8), 987-1010.

\bibitem{32} Zou, Qingyu, and Jing Bai. "Interest Points Detection in Image Based on Topology Features of Multi-level Complex Networks." Wireless Personal Communications: 1-11.


\bibitem{43} Gonzalez, Rafael C., and Richard E. Woods. "Digital image processing." (2002).


\bibitem{44} Bao, Paul, Lei Zhang, and Xiaolin Wu. "Canny edge detection enhancement by scale multiplication." IEEE transactions on pattern analysis and machine intelligence 27.9 (2005): 1485-1490.

\bibitem{46} Otsu, Nobuyuki. "A threshold selection method from gray-level histograms." IEEE transactions on systems, man, and cybernetics 9.1 (1979): 62-66.


\bibitem{48} Wani, M. Arif, and Bruce G. Batchelor. "Edge-region-based segmentation of range images." IEEE Transactions on Pattern Analysis and Machine Intelligence 16.3 (1994): 314-319.

\bibitem{42} Peng, Bo, Lei Zhang, and David Zhang. "A survey of graph theoretical approaches to image segmentation." Pattern Recognition 46.3 (2013): 1020-1038.


\bibitem{3} Wu, Zhenyu, and Richard Leahy. "An optimal graph theoretic approach to data clustering: Theory and its application to image segmentation." IEEE transactions on pattern analysis and machine intelligence 15.11 (1993): 1101-1113.

\bibitem{4} Felzenszwalb, Pedro F., and Daniel P. Huttenlocher. "Efficient graph-based image segmentation." International journal of computer vision 59.2 (2004): 167-181.


\bibitem{5} Shi, Jianbo, and Jitendra Malik. "Normalized cuts and image segmentation." IEEE Transactions on pattern analysis and machine intelligence 22.8 (2000): 888-905.


\bibitem{6} Li, Shijie, and Dapeng Oliver Wu. "Modularity-based image segmentation." IEEE Transactions on Circuits and Systems for Video Technology 25.4 (2015): 570-581.



\bibitem{7} Abin, Ahmad Ali, Farzane Mahdisoltani, and Hamid Beigy. "WISECODE: wise image segmentation based on community detection." The Imaging Science Journal 62.6 (2014): 327-336.



\bibitem{10} Linares, Oscar AC, et al. "Segmentation of large images based on super-pixels and community detection in graphs." arXiv preprint arXiv:1612.03705 (2016).



\bibitem{31} Newman, Mark. Networks: an introduction. Oxford university press, 2010.


\bibitem{33} Trémeau, Alain, and Philippe Colantoni. "Regions adjacency graph applied to color image segmentation." IEEE Transactions on image processing 9.4 (2000): 735-744.


\bibitem{8} Cigla, C., and Alatan, A. A. (2010, September). Efficient graph-based image segmentation via speeded-up turbo pixels. In Image Processing (ICIP), 2010 17th IEEE International Conference on (pp. 3013-3016). IEEE.

\bibitem{9} Comaniciu, Dorin, and Peter Meer. "Mean shift: A robust approach toward feature space analysis." IEEE Transactions on pattern analysis and machine intelligence 24.5 (2002): 603-619.

\bibitem{49} Cigla, Cevahir, and A. Aydın Alatan. "Efficient graph-based image segmentation via speeded-up turbo pixels." Image Processing (ICIP), 2010 17th IEEE International Conference on. IEEE, 2010.

\bibitem{47} Mori, G. (2005, October). Guiding model search using segmentation. In null (pp. 1417-1423). IEEE.


\bibitem{14} Puzicha, Jan, et al. "Empirical evaluation of dissimilarity measures for color and texture." Computer Vision, 1999. The Proceedings of the Seventh IEEE International Conference on. Vol. 2. IEEE, 1999.


\bibitem{13} Joyce, James M. "Kullback-leibler divergence." International encyclopedia of statistical science. Springer, Berlin, Heidelberg, 2011. 720-722.

\bibitem{30} Wright, W. D. "Color science, concepts and methods. Quantitative data and formulas." Physics Bulletin 18.10 (1967): 353.

\bibitem{11} Viola, Paul, and Michael J. Jones. "Robust real-time face detection." International journal of computer vision 57.2 (2004): 137-154.

\bibitem{12} Bernstein, Elliot Joel, and Yali Amit. "Part-based statistical models for object classification and detection." Computer Vision and Pattern Recognition, 2005. CVPR 2005. IEEE Computer Society Conference on. Vol. 2. IEEE, 2005.



\bibitem{26} R. Unnikrishnan, C. Pantofaru, and M. Hebert, "Toward objective evaluation of image segmentation algorithms," Pattern Analysis and Machine Intelligence, IEEE Transactions on, vol. 29, no. 6, pp. 929-944, 2007

\bibitem{27} Meila, M. (2005, August). Comparing clusterings: an axiomatic view. In Proceedings of the 22nd international conference on Machine learning (pp. 577-584). ACM.



\bibitem{34} Clauset, Aaron, Mark EJ Newman, and Cristopher Moore. "Finding community structure in very large networks." Physical review E 70.6 (2004): 066111.

\bibitem{35} Newman, Mark EJ. "Fast algorithm for detecting community structure in networks." Physical review E 69.6 (2004): 066133.

\bibitem{36} Delvenne, J-C., Sophia N. Yaliraki, and Mauricio Barahona. "Stability of graph communities across time scales." Proceedings of the National Academy of Sciences (2010).


\bibitem{17} Orman, G. K., Labatut, V., and Cherifi, H. (2011, June). Qualitative comparison of community detection algorithms. In International conference on digital information and communication technology and its applications (pp. 265-279). Springer Berlin Heidelberg.


\bibitem{18} Orman, G. K., Labatut, V., and Cherifi, H. (2011). On accuracy of community structure discovery algorithms. arXiv preprint arXiv:1112.4134.

\bibitem{19} Lancichinetti, A., and Fortunato, S. (2009). Community detection algorithms: a comparative analysis. Physical review E, 80(5), 056117.




\bibitem{20} Ronhovde, P., and Nussinov, Z. (2010). Local resolution-limit-free Potts model for community detection. Physical Review E, 81(4), 046114.


\bibitem{21} Rosvall, M., and Bergstrom, C. T. (2008). Maps of random walks on complex networks reveal community structure. Proceedings of the National Academy of Sciences, 105(4), 1118-1123.

\bibitem{22} Clauset, A., Newman, M. E., and Moore, C. (2004). Finding community structure in very large networks. Physical review E, 70(6), 066111.


\bibitem{23} Blondel, V. D., Guillaume, J. L., Lambiotte, R., and Lefebvre, E. (2008). Fast unfolding of communities in large networks. Journal of statistical mechanics: theory and experiment, 2008(10), P10008.














\bibitem{15} Sumengen, Baris, Luca Bertelli, and B. S. Manjunath. "Fast and adaptive pairwise similarities for graph cuts-based image segmentation." Computer Vision and Pattern Recognition Workshop, 2006. CVPRW'06. Conference on. IEEE, 2006.







\bibitem{25} P. Arbelaez, M. Maire, C. Fowlkes, and J. Malik, "Contour detection and hierarchical image segmentation," IEEE Transactions on Pattern Analysis and Machine Intelligence, vol. 33, no. 5, pp. 898-916, 2011.






\bibitem{28} A. Yang, J. Wright, Y. Ma, and S. Sastry, "Unsupervised segmentation  of natural images via lossy data compression," Computer Vision and Image Understanding, vol. 110, no. 2, pp. 212-225, 2008.

\bibitem{29} C. M. Christoudias, B. Georgescu, and P. Meer, "Synergism in low level vision," in Pattern Recognition, 2002. Proceedings. 16th International Conference on, vol. 4. IEEE, 2002, pp. 150-155.

\bibitem{30} Liu, Ye, et al. "From action to activity: sensor-based activity recognition." Neurocomputing 181 (2016): 108-115.


\bibitem{31} Liu, Ye, et al. "Action2Activity: Recognizing Complex Activities from Sensor Data." IJCAI. Vol. 2015. 2015.
























\end{thebibliography}
\end{document}